# AI DRIVEN HEALTH RECOMMENDER

K.Vignesh ,
Assistant Professor,
Department of Computer Science and Engineering
Kalasalingam Academy of Research and Education
vignesh.k@klu.ac.in

Chidura Sreenidhi
Department of Computer Science and Engineering
Kalasalingam Academy of Research and Education
sreenidhichidura@gmail.com

Boya Pranavi
Department of Computer Science and Engineering
Kalasalingam Academy of Research and Education
boyapranavi5@gmail.com

*Abstract*— ***In today's rapidly evolving healthcare landscape, timely and accurate access to health information is essential for improving patient outcomes, particularly in the early diagnosis and prevention of diseases. This project presents a novel machine learning-based system aimed at closing the gap between individuals and critical health insights. By employing predictive analytics, the system analyzes user-provided symptoms to deliver precise disease predictions and comprehensive health management advice, promoting preventive care and overall well-being.***

***The system leverages advanced algorithms such as Support Vector Machine (SVM), Gradient Boosting, and K-means clustering to provide detailed insights into predicted conditions, including associated symptoms, causes, and potential complications. Beyond disease prediction, it offers personalized recommendations tailored to each user's unique health profile. These suggestions encompass preventive actions, dietary advice, and personalized workout plans, empowering users to take control of their health proactively.***

***The system is implemented with an intuitive interface using Flask API, along with web technologies like HTML, CSS, and JavaScript, ensuring easy access for a wide range of users. By incorporating real-time analytics and personalized health solutions, this project has the potential to significantly enhance preventive care, reduce healthcare costs, and improve patient outcomes. Future developments will focus on expanding the system to integrate larger datasets and incorporating natural language processing (NLP) to better interpret unstructured inputs, enabling even more accurate and tailored health recommendations.***

## I. INTRODUCTION

The rapid evolution of technology has significantly impacted various sectors, particularly healthcare, where timely access to precise health information has become increasingly vital. In an era where early disease detection and personalized care are critical to improving patient outcomes, the use of artificial intelligence and machine learning offers a transformative opportunity for healthcare delivery. Machine learning algorithms have proven highly effective in analyzing vast datasets, recognizing patterns, and making precise predictions, thereby enhancing decision-making in clinical environments.

This paper presents an AI-powered health recommender system designed to provide users with timely disease predictions and personalized health advice based on their symptoms. The system empowers individuals to take a proactive role in managing their health by connecting them with critical healthcare insights. In addition to predicting potential diseases, the system provides information on associated symptoms, causes, and possible complications. Moreover, it offers customized recommendations, including precautionary steps, dietary advice, and workout plans tailored to the user's health profile, supporting overall well-being.

The system's functionality is powered by advanced machine learning algorithms such as Support Vector Machine (SVM), Gradient Boosting, and K-means clustering, which collectively enable accurate disease predictions and personalized recommendations. With a user-friendly interface built using Flask API, HTML, CSS, and JavaScript, the system is accessible to a broad user base. This research examines the system's potential to improve preventive care, lower healthcare costs, and enhance health outcomes, while also discussing future enhancements aimed at expanding predictive accuracy and personalization through the use of larger datasets and natural language processing.

## II. RELATED WORKS

Marouane Fethi Ferjani [1] proposed Common diseases 84.5% CNN detects features with high importance that renders better description of the disease, which enables it to accurately predict diseases with high complexity Kidney Disease SVM, NB. SVM considerably achieved higher accuracy than NB.Breast disease –RF, Bayesian Networks and SVM 95.85%.RF algorithm performed better than SVM because the former provides better estimates of information gained in each feature attribute. Since it scales well for large datasets and prefaces lower chances of variance and data overfitting

Faisal Rehman1,Osman Khalid [2] Introduced the food recommendation process using a variant of the Ant Colony Optimization (ACO) approach on a graph of foods.Diet-Right employs ACO, a constructive and population-based method inspired by ants' social behavior, recognized for solving combinatorial optimization problems. Food items are placed on nodes, forming a strongly connected graph with links associated with pheromone (τ) and heuristic information (η) values. The heuristic value (η) is initialized as the inverse of the squared sum of differences between all ingredients in food items.η is used to balance exploration and exploitation in the ACO algorithm, with its value ranging between 0 and 1.

Utkarsh Upadhyay,Aishwarya Srivastava [3] Emphasized the growing importance of maintaining a healthy lifestyle through a balanced diet, exercise, and avoiding poor food choices.Diet recommendation systems have emerged as tools offering personalized dietary guidance by leveraging technology, data science, and AI.These systems provide tailored advice based on individual factors like age, gender, activity level, and dietary preferences, addressing specific dietary restrictions and health conditions.Personalized dietary guidance marks a shift from the traditional one-size-fits-all approach, with the potential to prevent and manage diet-related diseases. The review article aims to examine the current state of diet recommendation systems, including their algorithms, applications, data sources, and challenges, and their impact on public health and general well-being. Guoguang Ronga,b, Arnaldo Mendez [4] proposed recent advancements in the application of AI in biomedicine, focusing on areas such as disease diagnostics, prediction, living assistance, biomedical information processing, and research.AI's role in biomedicine is growing due to both the continuous advancement of AI technologies and the complex nature of biomedical challenges that AI is well-suited to address.New AI capabilities are providing innovative solutions in biomedicine, while the evolving demands of biomedicine are driving further advancements in AI.The synergistic development between AI and biomedicine is expected to significantly advance both fields, ultimately improving the quality of life for people in need..P.K.SHANMUGAPRIYA,.MR.S.RAJKUMAR[5] improved efficiency and effectiveness in managing a fitness center by automating record-keeping and updating processes, replacing manual Excel file management.It offers increased efficiency, automation, accuracy, and a user-friendly interface. It also enhances communication capacity, maintenance, and cost reduction compared to existing systems.The system provides a user-friendly interface for data entry and management. Instructors can update fitness and diet plans based on individual user details, consulting with recognized trainers and dieticians.Users can retrieve data from the server using a key and generate daily reports, simplifying data access and management.The system is easy to design and implement, with minimal system requirements and compatibility with various configurations. Chintha Manoj Kumar,M Samyuel Sandeep reddy [6] They implemented model Using ML.They created a platform that recommends fitness videos based on the user Preference.The user can Join a weightloss plan based on their BMI. The proposed video recommendation system aims to enhance user engagement by combining implicit and explicit preferences such as viewing history and ratings. Existing systems like Fitness that Fits use a hybrid approach to recommend fitness videos based on user profiles and video attributes, offering diverse and engaging content. Suman bhoi,Mongli lee [7] They used Rule based ,Instance based and Longitudinal recommendations to convert Electronic Health Records[EHR] as graphs. The graphs are used to monitor the health of a patient in certain time periodThe EHR of a patient is represented as a sequence of visits, each described by multi-hot vectors for diagnosis, procedure, and medication.Drug co-occurrences and interactions are represented as graphs.The PREMIER system predicts personalized medications for a current visit by analyzing historical and current visit data. It uses neural attention models to weigh the importance of historical data and Graph Attention Networks to understand complex drug interaction. Abhishek, amitkumar bindal

[8] The dataset in `medicine.csv` includes 7 columns (5 character and 2 numeric), and analysis was performed using R libraries such as “recommender lab”, “ggplot”, `graph`, and `tidyr`. Key analyses included frequency counts of medicines, rating distributions, and heatmaps of ratings and drug similarity.The similarity between drugs was calculated using Cosine, Jaccard, and Pearson methods, with Cosine similarity being highlighted. The matrix was sparse due to missing ratings, so rows and columns with sums less than 60 were removed, resulting in an 885x3436 sparse matrix.They used ML approaches like collaborative filtering, content based filtering and Hybrid filtering to provide better results for recommending Certain Medicine name based on the Patient health condition. Arash Golibagh Mahyari, Peter Pirolli [9] The model is made with two inter-connected RNNs that takes exercise embedding and exogenous data as the input and recommends exercise activity and its probability of successful completion in the output.They used datasets of a mobile application names Dstress.proposed method is an innovative architecture of two inter-connected RNNs that takes exercise embedding and exogenous data as the input and recommends exercise activity and its probability of successful completion in the output. Rodrigo Zenun Franco, Computer Science University of Reading, UK [10] DEVELOP E-NUTRI SYSTEM: Created an online personalized nutrition advice system that uses FFQ data to provide dietary recommendations based on AHEI scores.USER EVALUATION: System positively evaluated by 163 participants for usability and acceptability.EXPERT FEEDBACK: Conduct dietary feedback survey with nutrition experts to validate and refine recommendations.UTILIZE POPULATION DATA: Enhance recommendations by incorporating data from national nutrition surveys.RANDOMIZED CONTROL TRIAL (RCT): Conduct an RCT to assess the effectiveness of personalized nutrition advice, comparing three groups with different levels of personalization

## III . METHODOLOGY

The ML-powered Health Adviser is a web application that utilizes machine learning to deliver personalized health recommendations based on the symptoms entered by users. Through an easy-to-use web interface, users can input their symptoms, which are then processed by the application, developed with the Flask web framework. The system uses a pre-trained machine learning model to predict the most likely disease and subsequently retrieves relevant information from a database. This information includes recommended treatments, medications, and dietary suggestions. The recommendations are then presented to the user in a clear and understandable format, helping them make informed decisions regarding their health.

**About Flask:**

Flask is a lightweight, flexible Python web framework that simplifies the development of web applications. It is user-friendly and efficient, offering essential functionalities like routing, templating, and request handling. Flask is ideal for both novice and experienced developers seeking to build scalable web applications without the complexity of larger frameworks.

**Recommender System**

A recommender system helps users discover relevant items, such as health suggestions, by analyzing user inputs or preferences. By using algorithms and data analysis, these systems predict the most appropriate recommendations for the user. They are widely utilized across various domains, from e-commerce to healthcare, to provide personalized suggestions based on user behavior and data.

COLLABORATIVE FILTERING**:**

In the context of medicine recommendation, collaborative filtering suggests treatments based on the symptoms by utilizing data from similar patients who have successfully undergone treatments for those symptoms.

CONTENT-BASED FILTERING:

This approach recommends medications by matching the patient's symptoms to drug profiles, relying on feature similarity between the symptoms and medications.

SUPPORT VECTOR MACHINE (SVM):

SVM is used to classify symptoms and predict the most effective medication by determining the optimal boundary between different treatment options.

K-NEAREST NEIGHBORS (KNN):

KNN suggests treatments by analyzing the medications given to patients with similar symptoms in the past.

**Deep Learning:**

Deep learning, through neural networks, models complex relationships in patient data, providing personalized medication recommendations based on symptom-treatment patterns learned over time.

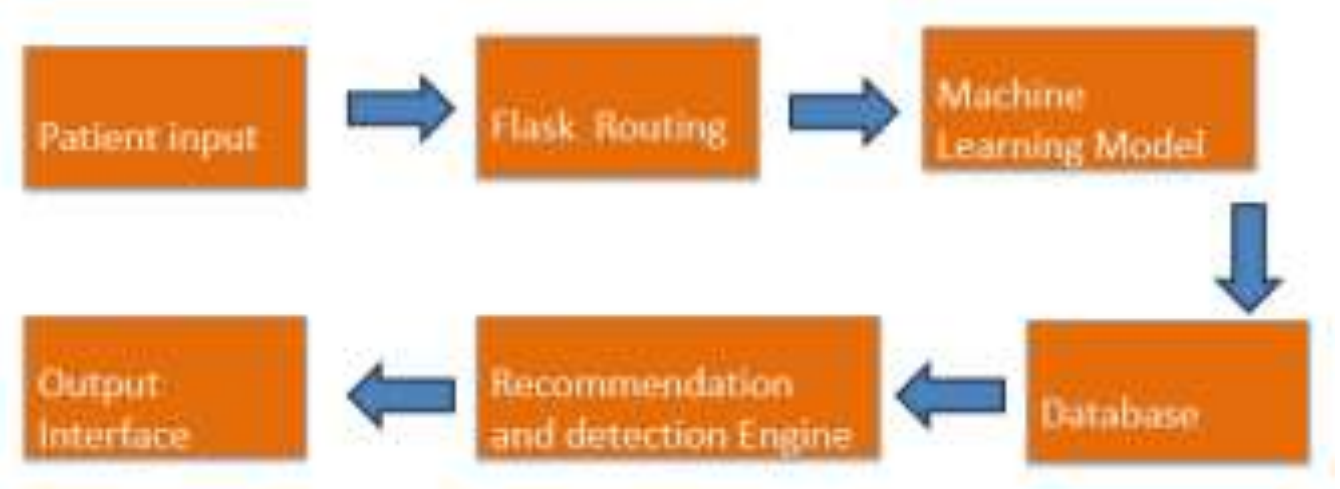

Figure 1 Block Diagram of Model

The system starts by allowing users to input symptoms they are experiencing or select them from a predefined list. The input is then sent to the backend, where the disease is detected, and appropriate medicines, precautions, and steps are recommended. The system processes the input using a machine learning model that has been trained with a comprehensive set of data. Once processed, the output is sent back to the user interface.

The ML-driven health adviser is essentially a recommendation system that follows several stages to produce tailored results. Since it must recommend various aspects such as disease names, medications, exercises, and precautions based on the symptoms provided, the dataset used must be accurate and well-prepared.

To achieve the best results, the system should utilize an algorithm that provides the highest accuracy and best fits the dataset. After selecting the optimal algorithm, it is used to train and test the model, ensuring the creation of an accurate recommender system. For a seamless user experience, Flask is utilized to connect the selected algorithm to the web server.

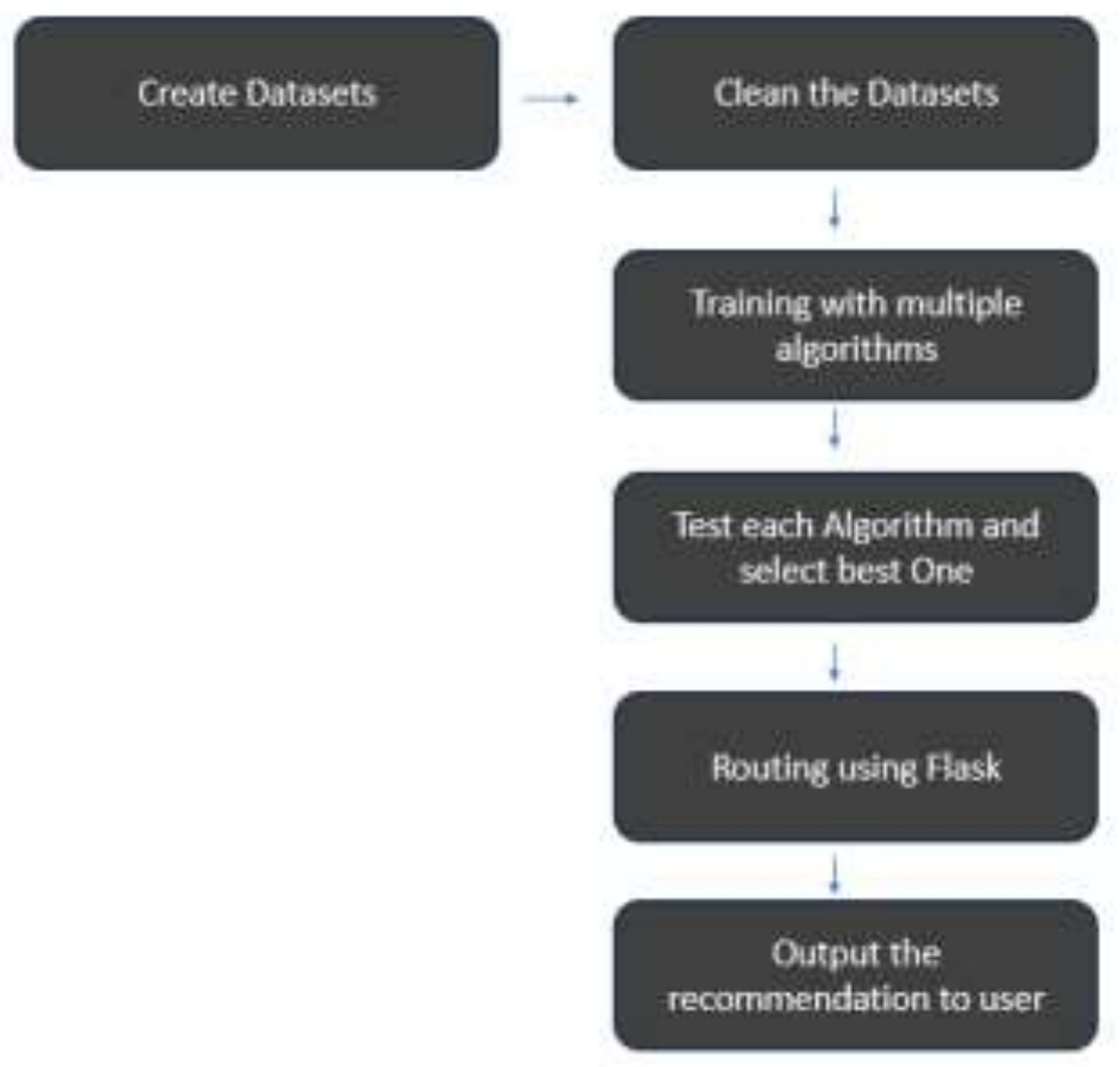

Figure 2 Flow Diagram of System

The flow of the system starts with users inputting their symptoms, which are then processed in the backend for disease detection The system also recommends appropriate medications, precautions, and lifestyle adjustments. The machine learning model, trained on relevant data, processes the input and returns the results to the user interface for easy access and understanding.

## IV.RESULT AND DISCUSSION

With the rise in similar symptoms across various individuals, accurately diagnosing diseases has become a challenge. However, by leveraging machine learning algorithms, we can detect diseases with considerable precision based on the symptoms provided. This technique can be applied in both web and mobile applications. Users input their symptoms, and the system analyzes this data to identify the most probable disease, subsequently offering recommendations for precautions and suitable diets.

A robust dataset, sourced from platforms such as Kaggle (consisting of 400 samples), is used to classify diseases into different categories. Based on the identified disease, the system suggests precautionary steps and dietary modifications. The data is split into 70% for training and 30% for testing, ensuring that the model is well-validated and capable of providing accurate predictions.

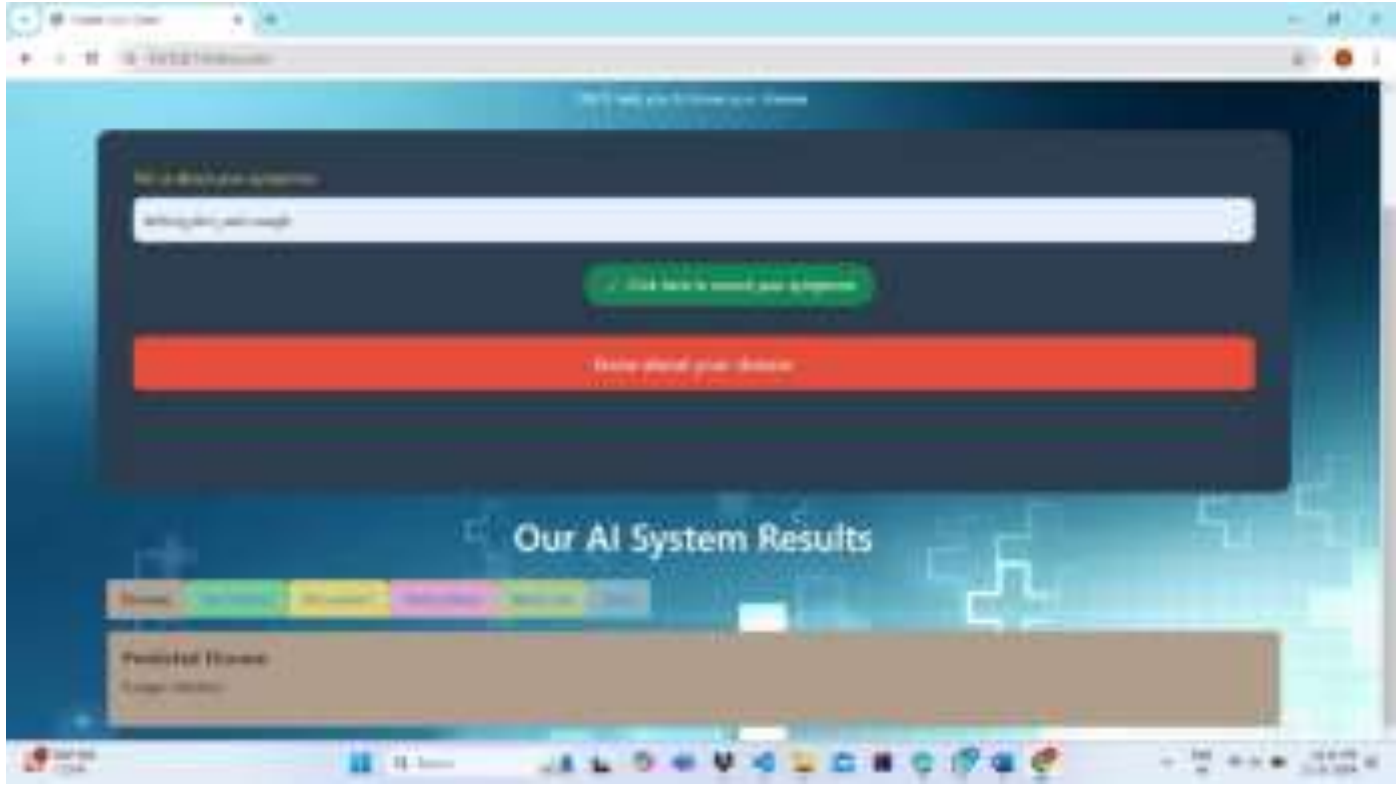

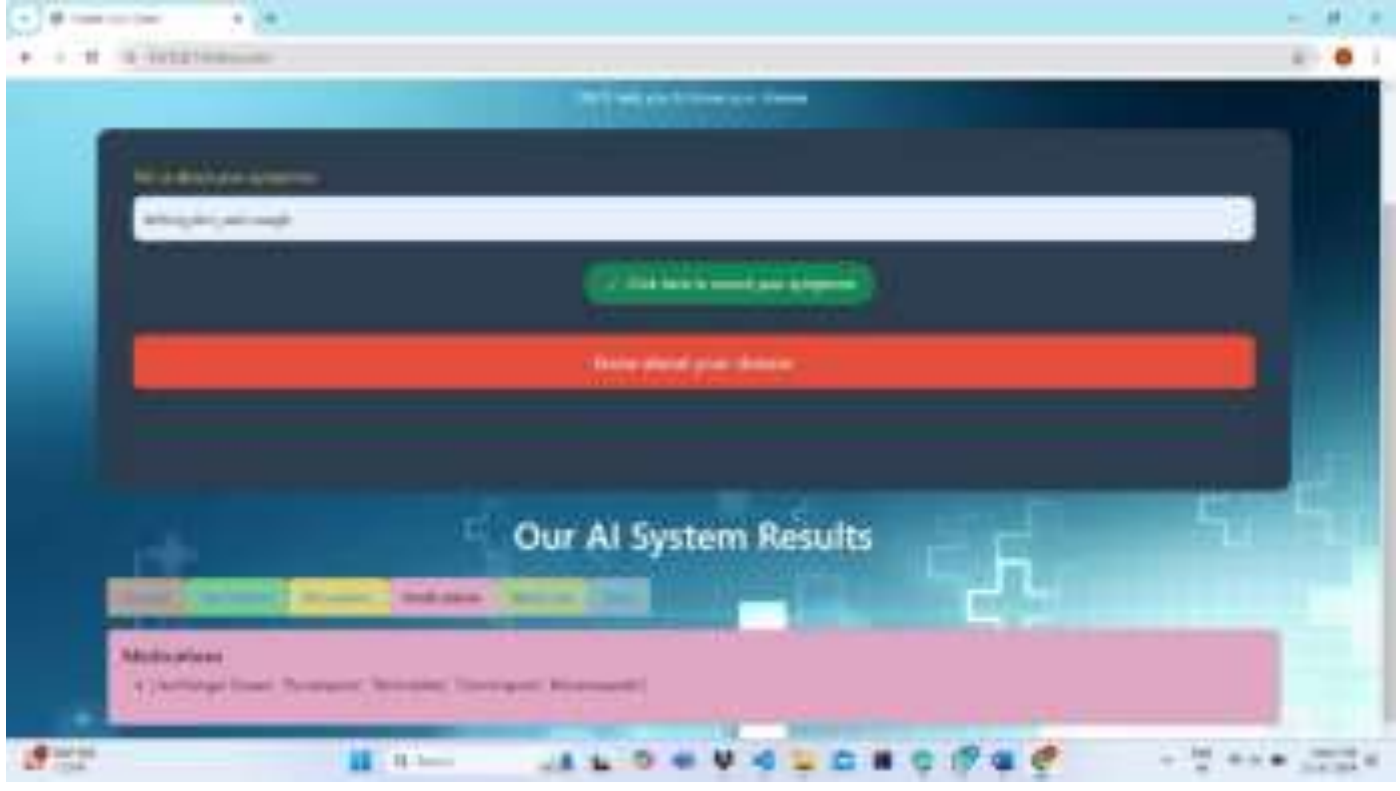

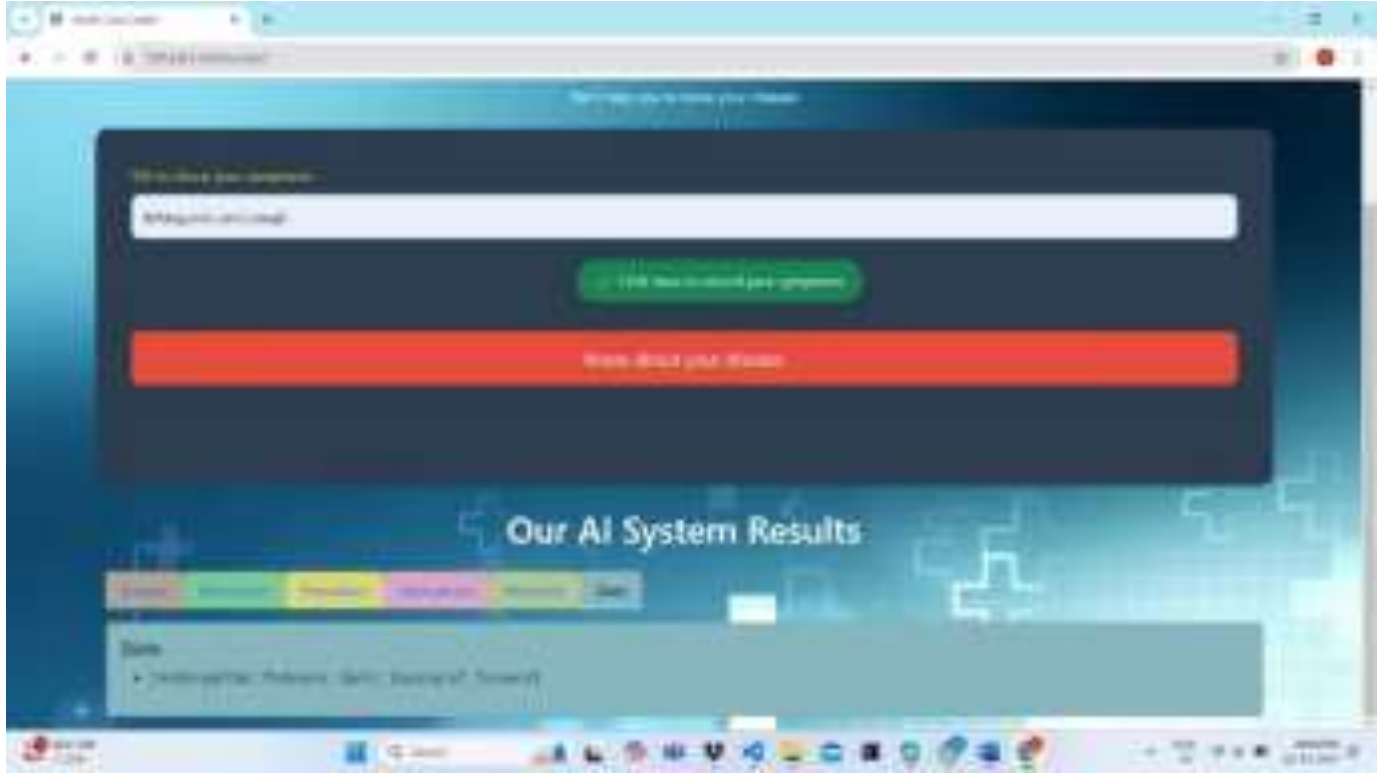

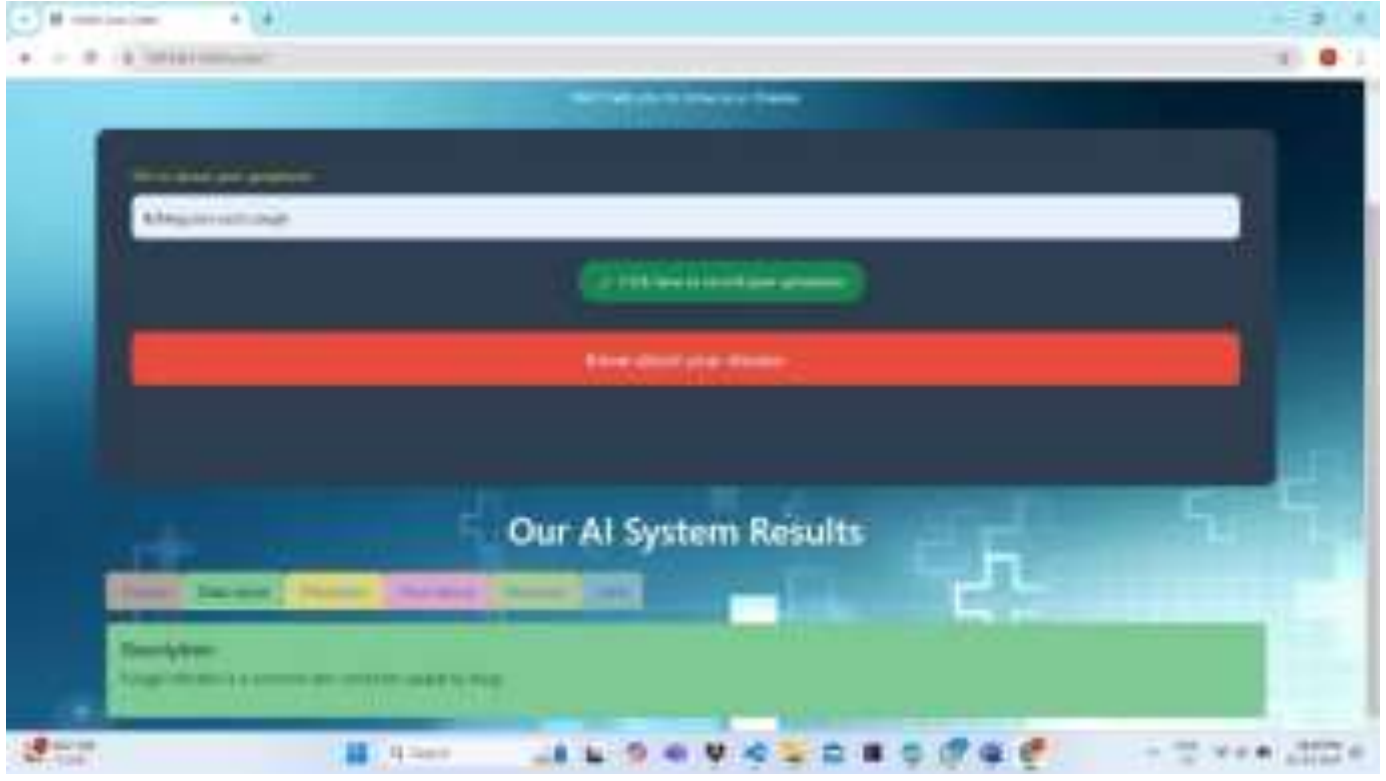

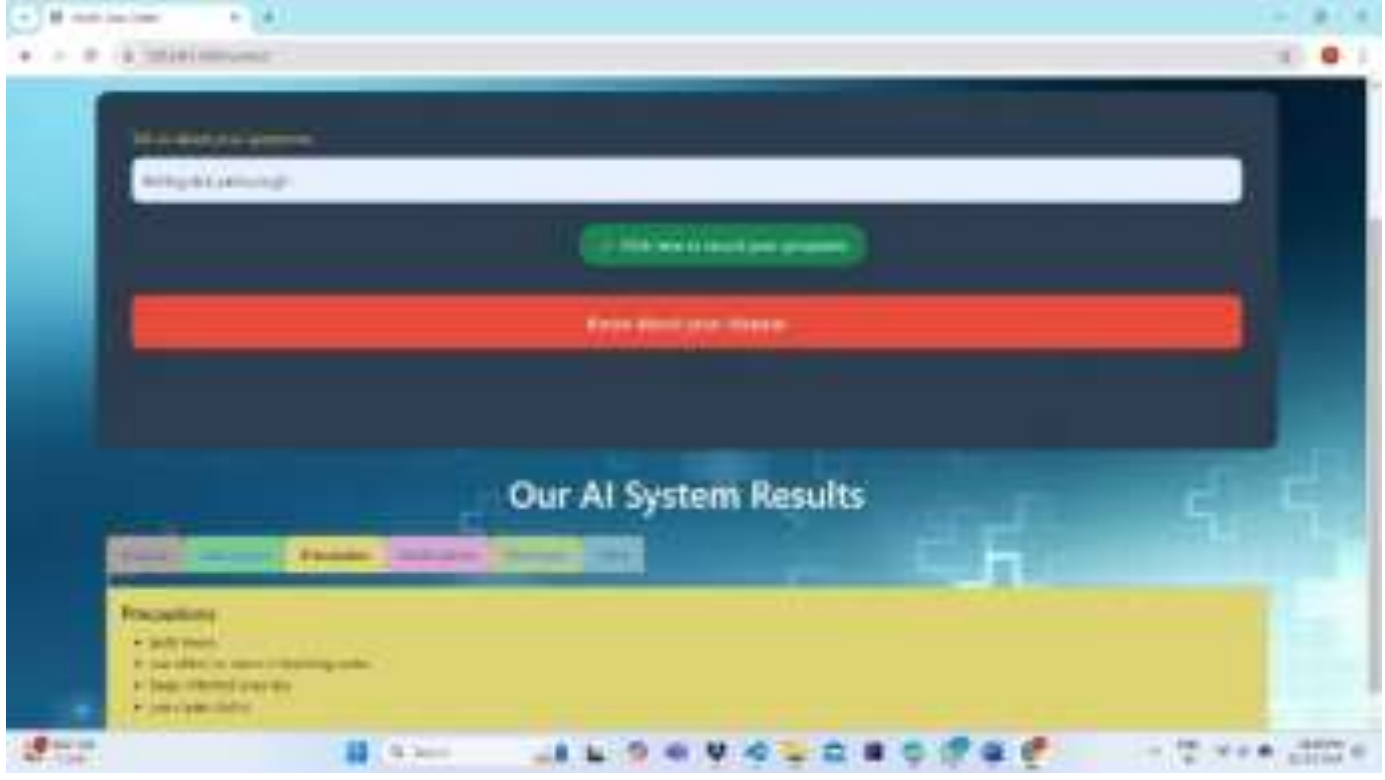

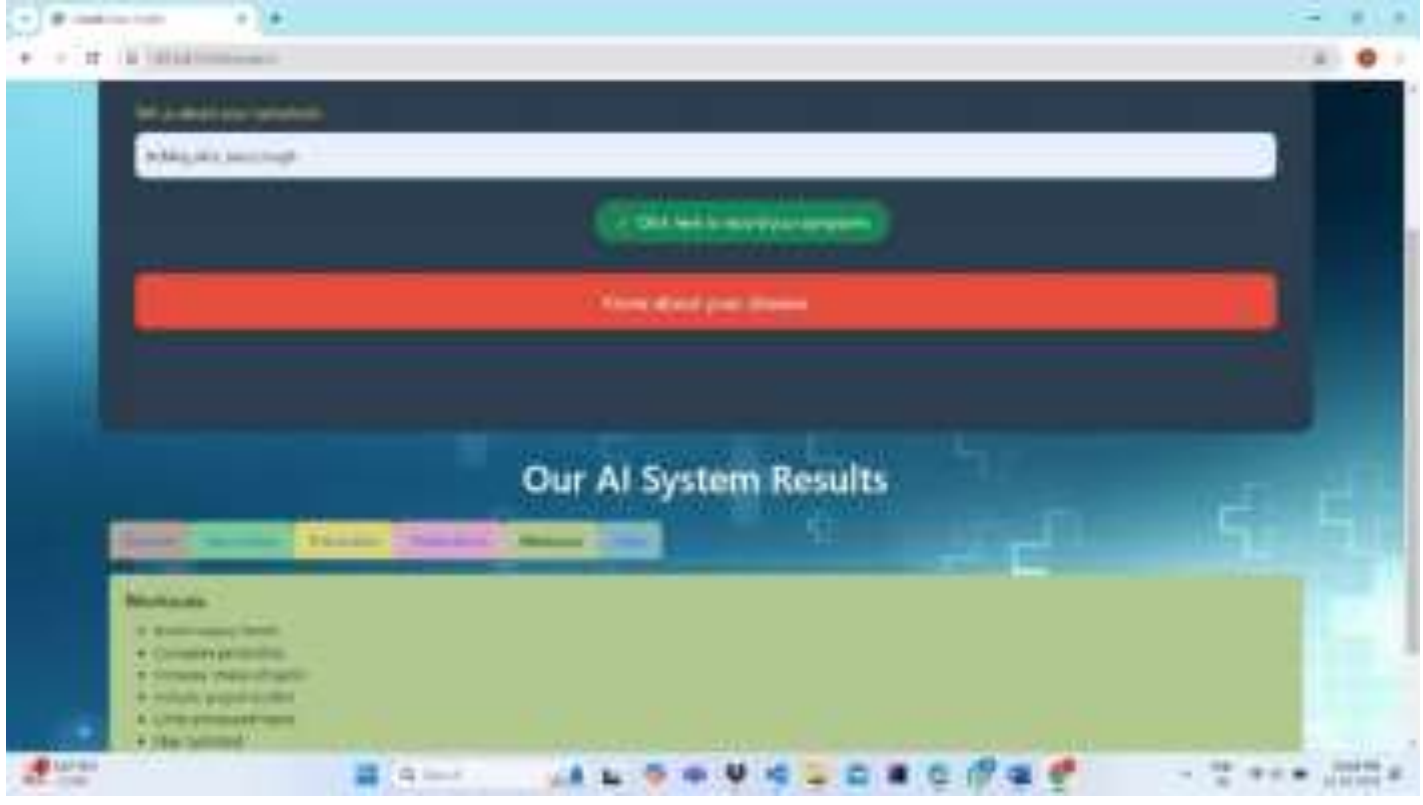

Once the machine learning model processes the symptoms entered by the user, it detects the corresponding disease and offers recommendations as illustrated in above Figures. This approach provides immediate health suggestions and helps monitor health trends across different age groups and demographics.

Gradient Boosting is a machine learning technique known for its high accuracy, particularly because it builds strong predictive models by combining several weak learners, such as decision trees. In the case of disease detection, Gradient Boosting is highly effective as it iteratively improves model performance by learning from previous mistakes at each step. Unlike other algorithms, Gradient Boosting is well-suited for complex and multi-dimensional data, making it a superior choice for detecting diseases where symptoms can vary widely between individuals. It excels at handling noisy and imbalanced datasets, common in healthcare applications, and effectively uncovers non-linear patterns in the data, leading to more accurate predictions.

By progressively minimizing errors, Gradient Boosting offers a more reliable prediction, making it ideal for disease detection systems that recommend relevant precautionary measures and dietary adjustments to users.

**Gradient Boosting Parameters:**

| Parameter | Description | Typical Rang |
|---|---|---|
| n_estimators | Specifies the number of boosting iterations or trees | 100-500 |
| learning_rate | Controls the rate of adjustment during each boosting step | 0.01 - 0.1 |
| max_depth | Sets the maximum depth for each tree | 3 - 7 |
| min_samples_split | Minimum number of samples needed to split an internal node | 2 - 10 |
| min_samples_leaf | Minimum number of samples required in the leaf nodes | 1 - 5 |
| subsample | Proportion of data used for training each tree | 0.6 - 1.0 |
| max_features | Maximum number of features to consider for each split | 'auto', 'sqrt', 'log2' |

Performance metrics:

Accuracy is used to measure the performance of the model

$$Acc = \frac{TP + TN}{TN + TP + FN + FP} \quad (1)$$

$$recall = \frac{TP}{TP+FN} \quad (2)$$

$$Precision = \frac{TP}{TP+FP} \quad (3)$$

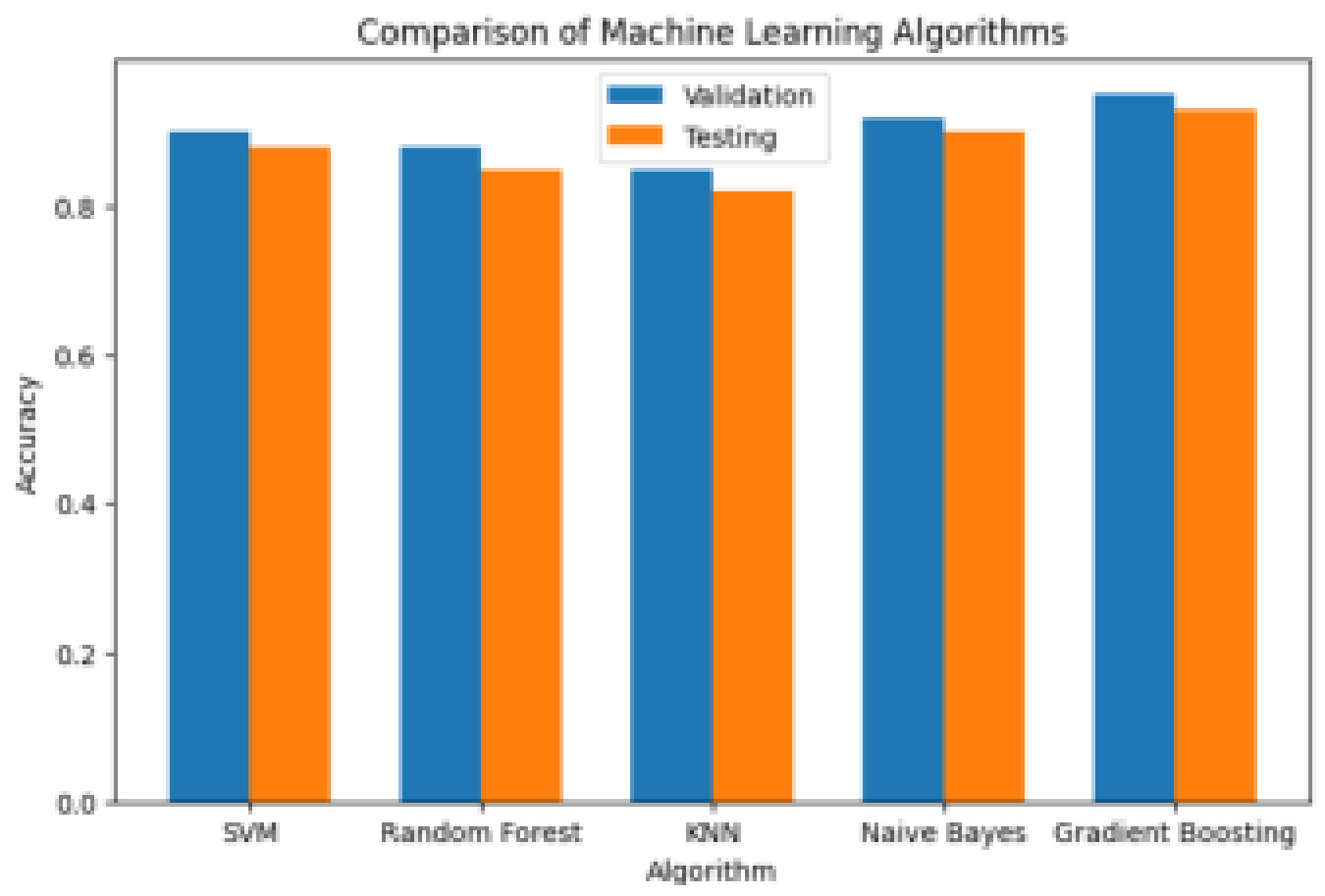

Figure: Comparison with existing techniques

From the Figure it is proved that our Gradient boosting method is more suitable for detecting the disease of any person by providing symptoms with 95% validation accuracy whereas Naïve Bayes yields 92%, Random Forest yields 88%, SVM yields 90% and KNN yields 85%. With the Flask implementation we proved that detecting the disease and preventions of the person based on the symptoms with Gradient boosting in efficient manner.

## V. CONCLUSION

In conclusion, the proposed machine learning-based system is poised to revolutionize preventative healthcare by offering a timely, accessible, and highly accurate disease prediction platform. By leveraging robust algorithms like Support Vector Machine (SVM), Gradient Boosting, and K-means clustering, the system effectively analyzes user-provided symptoms to generate precise predictions and actionable health insights. It not only identifies potential diseases but also provides comprehensive information on symptoms, causes, and possible complications, thereby equipping users with the knowledge to address health concerns early on. Moreover, the system's personalized recommendations—ranging from precautionary health measures to tailored dietary and workout plans—empower users to adopt healthier lifestyles and make informed decisions about their well-being. The intuitive user interface, developed using Flask API alongside HTML, CSS, and JavaScript, ensures that the platform is both user-friendly and accessible to a broad demographic. Looking ahead, the scalability of the system offers vast potential for future development. The integration of larger, more diverse datasets will enhance the accuracy and reliability of the predictions. Additionally, implementing natural language processing (NLP) will enable the system to better interpret and handle unstructured inputs, improving the user experience and further personalizing health advice. These enhancements will not only improve disease prediction but also help the system provide more precise recommendations, positioning it as a critical tool in reducing healthcare costs and improving patient outcomes through early intervention and preventative care. This system represents a meaningful step forward in the effort to make healthcare more proactive, personalized, and efficient.